# IMITATING U-NET ENHANCED AUTOENCODERS IN LATENT SPACE FOR IMPROVED PELVIC BONE SEGMENTATION IN MRI

*D. D. Pham[1], G. Dovletov[1], S. Warwas[2], S. Landgraeber[2], M. Jäger[2], and J. Pauli[1]*

[1]Intelligent Systems, Faculty of Engineering, University of Duisburg-Essen, Germany
[2]Department of Orthopedics and Trauma Surgery,
University Hospital Essen, University of Duisburg-Essen, Germany

**ABSTRACT**

We propose a 2D Encoder-Decoder based deep learning architecture for semantic segmentation, that imitates the encoder component of an U-net enhanced autoencoder in latent space and is trained in an end-to-end manner. Our proposed architecture is evaluated on the example of pelvic bone segmentation. A comparison to the standard U-net architecture shows promising improvements.

## 1. INTRODUCTION

Semantic segmentation is an important task in medical image processing, as it allows the generation of patient-specific 3D models for simulations and further diagnostics. Since manual segmentation is expensive and time consuming, research on fast automated approaches has sustained in the last decades. With the success of deep learning strategies on natural images, convolutional neural networks (CNNs) have also been introduced in the medical image processing domain and show state of the art performance. Especially Ronneberger et al.'s U-Net [1], proposed specifically for semantic segmentation tasks in medical images, has achieved a lot of attention, as this architecture yields impressive segmentation results in many medical applications compared to traditional approaches. Therefore, its general architecture has been modified in numerous ways in order to improve the segmentation quality even more, e.g. by extending the applicability to 3D volumes [2], modifying the loss function [3], or incorporating residuals [4] to name a few.

Recent research aims at incorporating shape priors into the segmentation process, as anatomical structures usually show only small shape variations. Ravishankar et al. [5] extend the U-Net by means of a pre-trained shape regularization autoencoder network, that is applied on the output of the U-Net to correct its output to a segmentation with feasible shape. Oktay et al. [6] propose a similar approach, in which also a pre-trained autoencoder is used to incorporate shape priors into the deep learning architecture. However, instead of merely correcting the initial segmentation output, they make use of the autoencoder's encoding component to regularize the weight adaptation process of a generic segmentation network during training, which is motivated by Girdhar et al.'s architecture for generating 3D representations of objects from 2D images [7]. Dalca et al. [8] go one step further by presenting an unsupervised learning scheme, employing pre-trained variational autoencoders, which also takes into consideration the shape priors.

In this work we also aim at incorporating anatomical priors, implicated by the ground truths, for an increased segmentation performance. However, instead of using pre-trained networks, we suggest a deep learning architecture for 2D images, that is end-to-end trainable.

## 2. METHODOLOGY

We propose an Imitating Encoder and Enhanced Decoder Network (IE$_2$D-Net) that imitates the encoding behavior of a Convolutional Autoencoder (CAE) in order to make use of the CAE's enhanced Decoder component. Our architecture consists of three major modules as depicted in Fig. 1. An U-Net subnetwork is used to extract relevant hierarchical features from the input image, in order to enhance the decoder component of a CAE, which forms the second module. The third module, namely the IE$_2$D-Net, comprises a combination of an imitating encoder, that aims at mirroring the generated feature vector of the CAE in latent space, and the decoder of the CAE, enhanced by the hierarchical U-net features. During training the U-Net and the imitating encoder receive a grayscale image as input, whereas the enhanced CAE is fed with the desired ground truth segmentation. However, during inference the encoder component of the CAE is omitted, relying on the imitation capabilities of the remaining imitating encoder.

### 2.1. Imitating the Encoder Component of the CAE

Convolutional Autoencoders are often used to find a compact representation of their input in latent space. The encoder component of the CAE maps the input to a low dimensional meaningful representation. The decoder component on the other hand is capable of reconstructing the original input from the compact representation. Motivated by Girdhar et al.'s TL-embedding network [7] we want to make use of this generative property to generate a feasible segmentation, given an arbitrary feature vector in latent space. Similar to Oktay et al.'s work [6], we use the CAE to find an embedding in latent space, that encodes the anatomical priors, given by the ground truth. We make use of an encoding component, consisting of subsequent convolutional and max-pooling layers, to imitate the feature vector, that would be generated by the CAE's encoder. This encoding component will be referred to as the *imitating encoder*. While the CAE's input and the desired output are the same, (i.e. the ground truth,) the imitating encoder's input is the corresponding medical (grayscale) image, e.g. a MRI or CT slice, as is depicted in Fig. 1. Therefore, the imitating encoder basically reduces the input image to the important anatomical priors in latent space, in order to be reconstructed to a segmentation by the CAE's decoder component. The imitation property of the imitating encoder is enforced by incorporating an imitation loss function that is only dependent on the kernel weights of the imitating encoder. Thus, minimizing this loss during training does not affect any weights of the other modules, and therefore solely encourages the encoder to mimic the feature vector in the CAE's latent space, given the corresponding medical (grayscale) image as input. A major difference to previous work lies in the end-to-end training approach in our proposal, in which both CAE and imitating encoder are trained at the same time, whereas the CAEs in Girdhar et al.'s and Oktay et al.'s approaches are pre-trained.

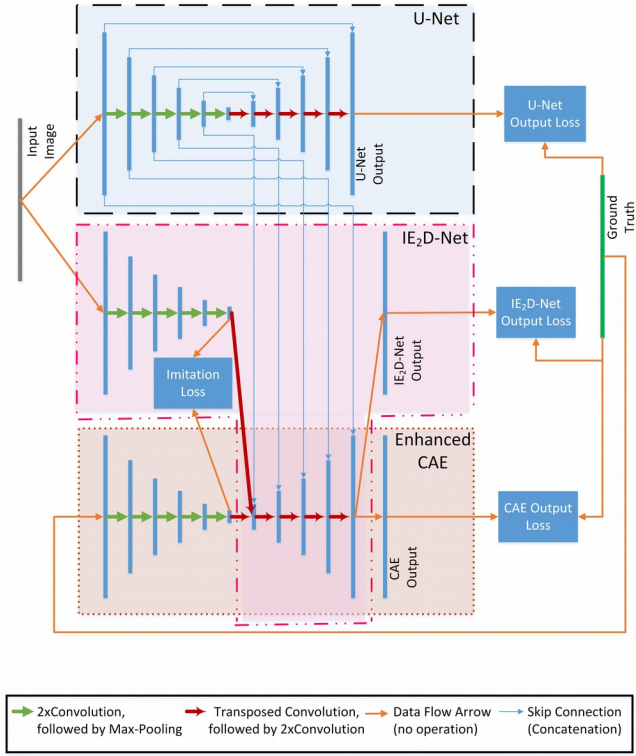

**Fig. 1**. Network Architecture. The architecture consists of three modules: the U-Net module (top), the CAE module (bottom) and the IE$_2$D-Net module (middle). For inference the encoder part of the CAE is omitted.

**2.2. Enhancing the Localization Capability of the CAE's Decoder**

For more robust decoder reconstructions, we employ skip connections, introduced in Long et al.'s Fully Convolutional Networks [9], which are also used in Ronneberger et al.'s U-Net [1] approach. The skip connections are used for better localization properties in the expansive path when increasing the resolution. This is achieved by combining features from each resolution level in the contracting path with the feature map in the corresponding resolution level of the expansive path. By introducing skip connections between the U-Net's contracting path and the corresponding resolution level of the CAE's decoder, we can make use of the learned hierarchical features from the U-Net to enhance the localization capability of the CAE's decoder. Since the U-Net is trained for segmentation purposes, whereas the imitating encoder is trained for imitation, we argue that the extracted features in the U-Net's contracting path is more suitable to enhance the CAE's decoder than the features extracted during the imitation process. This kind of enhancement remains plausible as long as minimizing the U-Net output loss function only affects the kernel weights within the U-Net component.

**2.3. Training and Inference**

During the training phase four different loss functions need to be minimized (Fig.1). The first loss function is the aforementioned imitation loss, that is only dependent on the kernel weights of the imitating encoder. A simple loss function would be the euclidean distance between the two vectors. The second loss function is the U-Net output loss, which only depends on the weights within the U-net module and measures the segmentation quality of the standard U-Net. The adaptation restriction to the U-Net weights ensures generation of hierarchical features in the contracting path, which are suitable for segmentation. The third loss function comprises of the CAE output loss, which only adapts the weights within the CAE module and judges the reconstruction quality of the CAE. Again, the restriction ensures the CAE to find a meaningful prior representation in latent space instead of depending on the features generated in the other modules. The last loss function consists of the IE$_2$D-Net output loss, i.e. a loss function that quantifies the segmentation quality of the combination of imitating encoder and the CAE's enhanced decoder. The minimization of this loss function is restricted to the adaptation of the imitating encoder weights and the CAE's decoder weights. For the last three loss functions the Dice Similarity Coefficient (DSC) or pixelwise cross entropy can be used. Since the CAE module and the IE$_2$D-Net module share the same decoder component, the corresponding loss functions need to be minimized successively. The successive optimization is also required for the IE$_2$D-Net output loss and the imitation loss, since both are dependent on the imitating encoder weights.

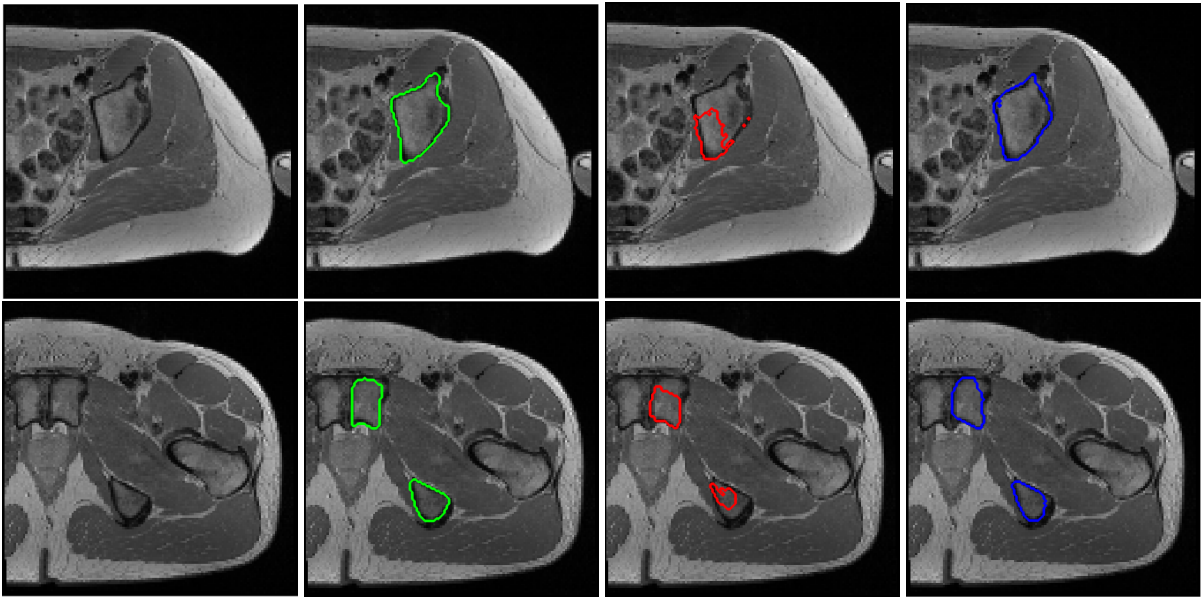

**Fig. 2**. Exemplary results. The first column shows the exemplary axial MRI slice. The ground truth (green) is depicted in the second column, the U-Net output (red) in the third, and the IE$_2$D-Net output (blue) in the last column.

During inference the encoding component of the CAE is replaced with the imitating encoder. Therefore, the U-Net module extracts relevant features in each resolution level in its contracting path and at the same time the imitating encoder projects the input into a compact representation for priors in latent space. This representation is then decoded by the decoder component, enhanced by the extracted U-Net features, resulting in the final segmentation.

## 3. EXPERIMENTS

### 3.1. Data

We applied our proposed architecture on the example of the segmentation of the pelvic bones in MR images and compared our results to the segmentations achieved solely by the U-Net module, i.e. a standard U-Net architecture. We used data sets, approved by our institution's ethics committee, consisting of eight T1-weighted MRI volumes comprising six different patients. For two of these patients MRI scans were obtained before and after a surgical procedure. We denote the data sets as $P1, \ldots, P6$ and mark the post operative data sets as $\overline{P}1$ and $\overline{P}2$. Each volume consists of 40 to 44 axial slices. The experiments were conducted in a leave-one-out cross validation manner, in which one data set is kept for testing, and the remaining data sets were used for training. In case of P1, $\overline{P}1$, P2, and $\overline{P}2$ only those data sets were used for training, which do correspond to the same patient. P6 was used as validation data set to monitor the training process. We increased the number of slices by augmenting the training data by means of rotation and translation.

### 3.2. Implementation Details

We resized the MRI slices to an input size of $128 \times 128$ and used an encoder resolution depth of 5 for the U-Net's contracting path, the imitating encoder and the CAE's encoder. For each resolution level we used two convolutional layers, succeeded by one max-pooling layer on the contracting/encoding side. On the expansive/decoding side we used one transposed convolution layer, followed by two convolutional layers for each resolution level in each module, as depicted in Fig. 1.

Starting with 16 kernels for each convolutional layer in the first resolution level, we doubled the number of kernels for each resolution level on the contracting/encoding side and halved the number of kernels on the expansive/decoding side. We used a kernel size of $10 \times 10$ for every convolutional layer and $2 \times 2$ max-pooling.

For the imitation loss we used the euclidean distance and for the remaining loss functions we subtracted the Dice Similarity Coefficient (DSC) from 1. We implemented our architecture in Tensorflow and ran our experiments on a GTX 1080 ti GPU.

## 3.3. Results

Our proposed architecture yields promising results, compared to the U-Net segmentations. Fig. 2 shows exemplary MRI slices from two different patients with the desired segmentation (green), the U-Net output (red) and our result (blue). While the U-Net apparently has difficulties to encapsulate the bone contour, our approach shows superior results in these examples. Table 1 shows the achieved DSC values from the U-Net and our proposed IE$_2$D-Net in comparison. Although for P5 the U-Net shows better performance, our approach shows similar and even better results than the U-Net in the remaining data sets, leading to an average DSC of $73.45 \pm 5.93$ compared to an average DSC of $69.94 \pm 7.44$ for the U-Net. We assume that the U-Net performs better on P5 because the difference in bone intensity and fat intensity is smaller than in the remaining data sets. Thus, our architecture might profit from intensity augmentations during training. Additionally, a fusion of U-Net module output and IE$_2$D-Net output may lead to improved segmentation results.

|  | U-Net | IE$_2$D-Net |
|---|---|---|
| P1 | 63.42 | **66.32** |
| $\bar{P}$1 | 57.19 | **67.15** |
| P2 | 67.84 | **74.66** |
| $\bar{P}$2 | 76.85 | **84.56** |
| P3 | **80.34** | 76.5 |
| P4 | 78.85 | **78.86** |
| P5 | 65.25 | **66.13** |
| ∅ | 69.94± 7.44 | 73.45± 5.93 |

**Table 1**. Resulting average DSCs of the U-Net and IE$_2$D-Net for each data set.

## 4. CONCLUSION

In this work we present a 2D deep learning architecture that shows promising results on the segmentation of pelvic bones. Altogether our contribution is two-fold. We introduced the enhancement of the decoder component of an CAE by means of U-Net features and presented an end-to-end learning scheme for the incorporation of priors. Our evaluation shows promising segmentation results compared to the U-Net. Performance improvements might be gained by means of additional intensity augmentation during training and fusion of the U-Net module output with the IE$_2$D-Net output. In the future, we intend to extend our approach to 3D and introduce variational autoencoders to increase the robustness of the decoder.